\documentclass{article}
\usepackage{spconf,amsmath}
\usepackage[pdftex]{graphicx}
\usepackage{subcaption}
\usepackage{amssymb}
\usepackage{comment}

\newcommand{\R}{\mathbb{R}}

\title{LAYER-WISE INTERPRETATION OF DEEP NEURAL NETWORKS \\USING IDENTITY INITIALIZATION}
\name{Shohei Kubota$^{\dagger \diamondsuit}$, Hideaki Hayashi$^{\dagger \diamondsuit}$, Tomohiro Hayase$^{\star}$, and Seiichi Uchida$^{\dagger}$ \thanks{This work was supported by JST ACT-I grant number JPMJPR18UO, JST ACT-X grant number JPMJAX190N, and JSPS KAKENHI Grant Number JP17H06100. ${}^\diamondsuit$ Contributed equally. \textcopyright 2021 IEEE. Personal use of this material is permitted. Permission from IEEE must be obtained for all other uses, in any current or future media, including reprinting/republishing this material for advertising or promotional purposes, creating new collective works, for resale or redistribution to servers or lists, or reuse of any copyrighted component of this work in other works.}}
\address{$^{\dagger}$ Kyushu University, Japan, $^{\star}$ Fujitsu Laboratories, Japan}
\begin{document}
\setlength{\abovedisplayskip}{3pt} 
\setlength{\belowdisplayskip}{3pt} 
%
\maketitle

\begin{abstract}
The interpretability of neural networks (NNs) is a challenging but essential topic for transparency in the decision-making process using machine learning. One of the reasons for the lack of interpretability is random weight initialization, where the input is randomly embedded into a different feature space in each layer. In this paper, we propose an interpretation method for a deep multilayer perceptron, which is the most general architecture of NNs, based on identity initialization (namely, initialization using identity matrices). The proposed method allows us to analyze the contribution of each neuron to classification and class likelihood in each hidden layer. As a property of the identity-initialized perceptron, the weight matrices remain near the identity matrices even after learning. This property enables us to treat the change of features from the input to each hidden layer as the contribution to classification. Furthermore, we can separate the output of each hidden layer into a contribution map that depicts the contribution to classification and class likelihood, by adding extra dimensions to each layer according to the number of classes, thereby allowing the calculation of the recognition accuracy in each layer and thus revealing the roles of independent layers, such as feature extraction and classification.
\end{abstract}
\begin{keywords}
Multilayer perceptron, explainable AI, identity initialization
\end{keywords}
{\allowdisplaybreaks
\vspace{-2mm}
\section{Introduction}
\label{sec:intro}
\vspace{-3mm}
Neural networks (NNs), including deep multilayer perceptrons (MLPs), have been used in various applications, such as signal processing~\cite{zhang2019robust, braun2019parameter,zhang2019investigation} and image classification~\cite{bei2020combining, srinidhi2020validational, shankar2020hyper}. However, NNs consist of nonlinear functions and numerous parameters and thus lack interpretability. In other words, they are regarded as black boxes. Methods that explain the inner process of NNs, also referred to as explainable AI~\cite{david2019explainable_}, are therefore required, which leads to an investigation of the causes of misclassification and verification of adherence to ethical standards. \par

One of the reasons for the lack of interpretability is the use of random numbers for weight initialization~\cite{glorot2010Understanding,He2015Delving, Klambauer2017Advances}. It happens because the input is randomly embedded into a different feature space in each layer. Many studies have been conducted to interpret the inner process by calculating the contribution of each neuron to the decision~\cite{iwana2019explaining,selvaraju2017grad,zhou2016learning,montavon2017explaining,bach2015pixel}. Such methods require backward calculations such as backpropagation, which is computationally expensive. It is also difficult to calculate the class likelihood for each layer. \par

For better interpretability, we focus on identity initialization. Identity initialization is a non-random weight initialization method in which the weight matrices of an MLP consisting of fixed-width hidden layers are initialized with scaled identity matrices. An MLP is the most general architecture of NNs, and analyzing the MLP will lead to future applications in other architectures. \par

Identity initialization demonstrates its potential for interpretability when the network becomes deeper. It is experimentally known that local optima of weights in an over-parameterized NN exist around the initial random weights~\cite{li2018learning, Oymak2019overparam}. The learned weights of an identity-initialized NN are also expected to be close to the identity matrix because deeply stacked layers make the transformation in each layer minute. In this case, the output of each hidden layer has a slight variation from the layer's input, and the amount of variation can be regarded as the contribution of each neuron. If we can make the identity-initialized MLP deeper without causing the gradient vanishing/exploding problem, we can interpret its inner process. \par

In this paper, we propose an interpretation method for a deep MLP based on identity initialization. To realize a deep identity-initialized MLP, we first perform a theoretical analysis of the identity initialization. Under the assumption that the weight matrices of all hidden layers are the identity matrix, we demonstrate how to propagate forward/backward signals between the input and last layers without vanishing or exploding gradients, even with a huge number of layers. We also propose a network structure that can further enhance interpretability. In the proposed structure, we can separate internal features into contribution maps for classification and class likelihoods, thereby enabling us to calculate classification accuracy in each layer, thus revealing the layer-wise discriminability. \par

The main contributions of this study are as follows:
    1) We theoretically demonstrate that forward/backward propagation does not vanish/explode in an identity-initialized MLP; 
    2)  We experimentally demonstrate that the learned weights of an identity-initialized MLP remain near the identity matrix; 
    3)  We propose an MLP structure that allows us to separate each hidden layer's output into a contribution map and class likelihood.

\vspace{-4mm}
\section{Identity initialization and \\Signal Propagation}
\vspace{-3mm}
\begin{figure}[t]
	\centering
	\includegraphics[width=0.9 \hsize]{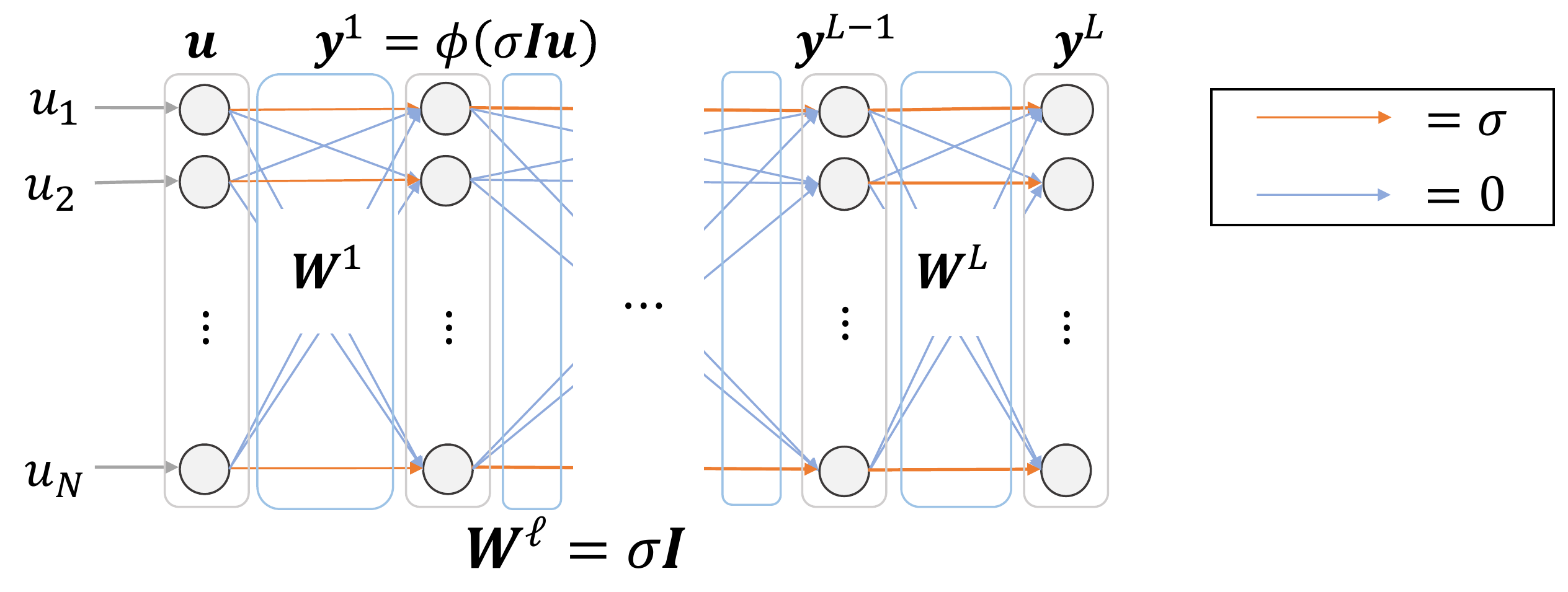}
    \caption{Identity initialization of a deep MLP. The weight matrix of the $\ell$-th layer is initialized with the identity matrix as $W^\ell = \sigma I$. The orange and blue arrows represent connections with weights of $\sigma$ and 0, respectively.}
	\label{Fig:Identity_initialization}
\end{figure}
\begin{figure*}[t]
	\centering
	\includegraphics[width=0.85 \hsize]{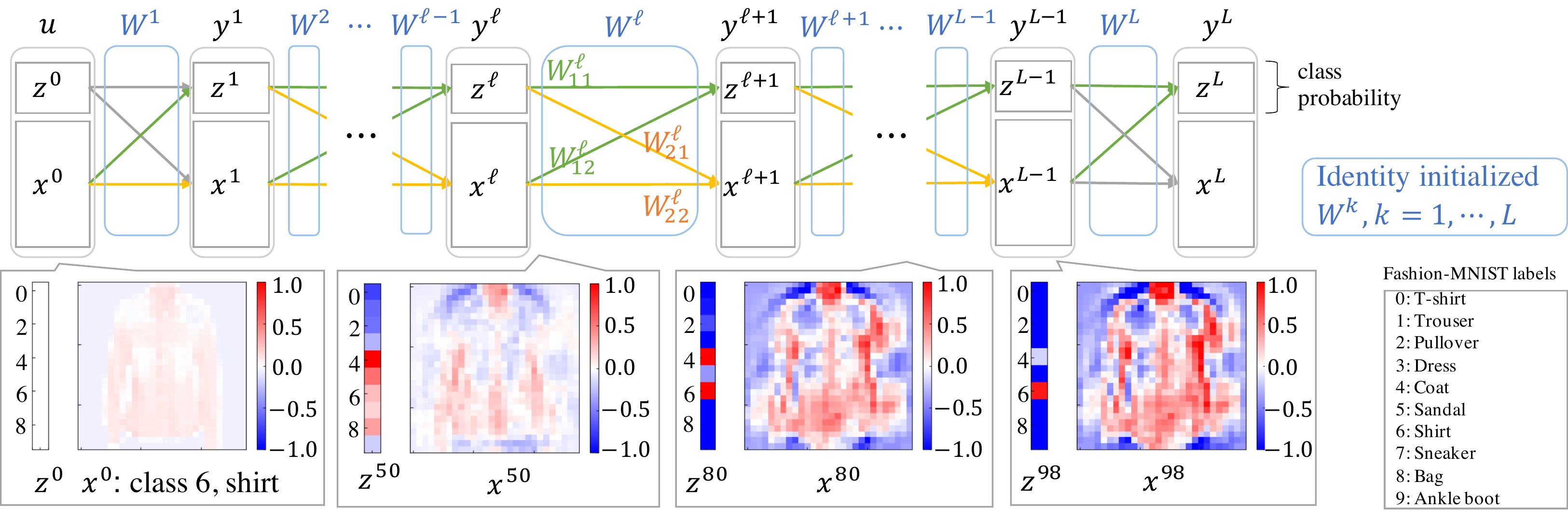}
    \caption{Diagrammatic representation of the proposed interpretable MLP structure. Each hidden layer consists of two sub-vectors $z^\ell$ and $x^\ell$, and they involve the class likelihood and contribution map, respectively. Note that $x^\ell$ is an $N$ vector for the $\sqrt{N} \times \sqrt{N}$ image input, and we visualize $x^\ell$ as an $\sqrt{N} \times \sqrt{N}$ image for better visibility.}
	\label{Fig:Network}
\end{figure*}
Here, we introduce a deep MLP with identity initialization. For interpretability, we make the MLP quite deep to make the weight matrices close to their initial state. We also discuss how to set the MLP. 
\vspace{-3mm}
\subsection{Identity initialization}
\vspace{-1mm}
Consider an MLP of $L$ layers with width $N$ and weight matrices $W^\ell \in \R^{N \times N}$ $(\ell = 1, \ldots, L)$, as shown in Fig.~\ref{Fig:Identity_initialization}. Each weight matrix is initialized as follows:
\begin{equation}
    \label{eq:identity_init}
    W^\ell = \sigma I,
\end{equation}
where $\sigma$ is a constant positive value and $I \in \R^{N \times N}$ is an identity matrix. The output of the $\ell$-th layer, $y^\ell$, is defined as
\begin{equation}
    \label{eq:forward_propagation}
    y^\ell = \phi(h^\ell), \quad h^\ell = W^\ell y^{\ell - 1},
\end{equation}
where $\phi$ is an activation function and $h^\ell$ is a pre-activation vector. The input to the MLP is given as $u$. The entries of $u$ are normalized to have a mean of zero and variance of $q^0$. \par

\vspace{-3mm}
\subsection{Signal propagation and dynamical isometry}
\vspace{-1mm}
\label{sec:dynamical isometry}
We show the theoretical result of how a deep MLP with identity initialization propagates a signal without exploding or vanishing. If the network is quite deep, naive settings (i.e., general initialization and activation functions) cause the vanishing/exploding gradients of networks and make it difficult to reduce the training error. \par

The \emph{dynamical isometry}~\cite{dynamical_isometry, Pennington2018emergence} is introduced to prevent the vanishing/exploding gradients. The dynamical isometry means that all singular values of the input-output Jacobian $J$ (or, equivalently, eigenvalues of $JJ^\top$) of the network concentrate around $1$. If we can set the MLP in such a way that it satisfies the dynamical isometry asymptotically, we can take a large value of $L$. \par

Now, the input-output Jacobian of the MLP is given by
\begin{equation}
    \label{eq:Jacobian}
    J = \frac{\partial y^L}{\partial u} = D^1 W^1 D^2 W^2 \cdots D^L W^L, 
\end{equation}
where $D^\ell$ is a diagonal matrix with entries $D^\ell_{i,j} = \phi^{'}(h^\ell_{i}) \delta_{ij}$ and thus depends on $h^\ell$. The entries of $h^\ell$ follow the distribution $\rho^\ell$, which has a zero mean and variance $q^\ell$. Owing to the recursive nature of \eqref{eq:forward_propagation}, $q^\ell$ has a fixed point that satisfies $q^1 = q^L$ when $q^0 = q^*$. The existence of $q^*$ guarantees the initial forward propagation without vanishing or exploding, and $q^*$ is also required for deriving the eigenvalue distribution. Specifically, when $\phi$ is a hard-tanh function $\phi(x) = \mathrm{ReLU}(x + 1) - \mathrm{ReLU}(x - 1) -1$ and $x^0$ follows a zero-mean Gaussian with variance $q^*$, $q^*$ obeys
\begin{equation}
\label{identity_forward}
    q^* = \frac{\beta-2}{2} {\rm erf}\left(\frac{1}{\beta} \right) - \sqrt{\frac{\beta}{\pi}} \exp{\left(-\frac{1}{\beta}\right)} + 1, 
\end{equation}
where $\beta=2\sigma^{2(\ell - 1)} q^*$, and $\rm erf$ is the error function\footnote{Eq.~\eqref{identity_forward} is derived from the variance of the $L$-th layer's output distribution, which is calculated by applying~\eqref{eq:forward_propagation} to a Gaussian distribution $L$ times.}. \par

Substituting~\eqref{eq:identity_init} for~\eqref{eq:Jacobian}, the Jacobian becomes a diagonal matrix $J = \sigma^L \prod_{\ell=1}^{L} D^\ell$. Hence the eigenvalue distribution is calculated by the product of the diagonal parts of $JJ^\top$:
\begin{equation}
    \mu_{JJ^\top} = \sigma^{2L} \prod_{\ell=1}^{L} \phi^{'}(\rho^\ell)^2.
\end{equation}
Using the same settings as \eqref{identity_forward}, $\mu_{JJ^\top}$ is expressed as follows:
\begin{flalign} \label{identity_backward}
    \mu_{JJ^\top}(dx) \!=\!\begin{cases}
        \alpha^1 \delta_{\sigma^{2L}}(dx) \!+\! (1 \!-\! \alpha^1) \delta_0(dx) & \!\!\!\!(\sigma \!\leq\! 1), \\
        \alpha^L \delta_{\sigma^{2L}}(dx) \!+\! (1 \!-\! \alpha^L) \delta_0(dx) & \!\!\!\!(\sigma \!>\! 1), 
    \end{cases}
\end{flalign}
where $\alpha^\ell = {\rm erf}\left(1/\sigma^\ell \sqrt{2 q^*}\right)$, and $\delta_p$ is a Dirac measure at $p \in \R$. \par

From \eqref{identity_forward} and \eqref{identity_backward}, by appropriately setting $\sigma$ and $q^*$, we obtain singular values concentrated around 1, thereby preventing the vanishing/exploding gradients. In practice, $\sigma$ and $q^*$ are chosen from the numerical solutions of~\eqref{identity_forward} such that approximately 80\% of the singular values concentrate 1.
\vspace{-5mm}
\section{Interpretable MLP based on \\identity initialization}
\label{sec:interpretable}
\vspace{-3mm}
We propose an interpretable deep MLP structure for classification tasks based on identity initialization. The proposed method allows for the visualization of the contribution to classification and encoding of the class likelihood in each layer. \par

\vspace{-3mm}
\subsection{Network structure enhancing interpretability}
\vspace{-1mm}
\label{enhancing_interpretability}
The proposed interpretable network consists of $L$ layers. In the input layer, instead of directly inputting a data sample $x^0 \in \R^N$, we expand the dimension by concatenating a zero vector $z^0 = (0 \ \cdots \ 0)^\top \in \R^C$, where $C$ is the number of classes, with $x^0$; namely, the input to the network is represented as $u = (z^0 \ x^0 )^\top \in \mathbb{R}^{C + N}$,
where the entries of $x^0$ are normalized to have a zero mean and variance of $q^*$. Each hidden layer has a width of $C+N$, and its weight matrix $W^\ell \in \R^{(C+N)\times(C+N)}$ is initialized with an identity matrix. The teacher vector is given as a $C$-dimensional one-hot vector, and the cross-entropy loss is calculated based on $z^L$ and the teacher vector during training. \par

Owing to this structure, the proposed network is diagrammatically represented, as shown in Fig.~\ref{Fig:Network}. In this representation, we regard each hidden layer output $y^\ell$ to be composed of $z^\ell \in \R^C$ and $x^\ell \in \R^N$, as with the input layer. The internal features $z^\ell$ and $x^\ell$ are interpretable as layer-wise class likelihoods and contribution maps, respectively. The $C$-dimensional vector $z^L$ gives the final class likelihood. \par

The weight matrix of each hidden layer can be divided into four sub-matrices
\begin{equation}
W^\ell = \left(
    \begin{array}{cc}
      W_{11}^\ell & W_{12}^\ell \\
      W_{21}^\ell & W_{22}^\ell 
    \end{array}
  \right),
\end{equation}
where $W_{11}^\ell \in \mathbb{R}^{C\times C}$ and $W_{22} \in \mathbb{R}^{N \times N}$.
Then, $y^{\ell}$ is represented as
\begin{equation}
\label{eq:forward_with_submatrix}
    y^\ell = \left(
    \begin{array}{c}
      z^\ell \\
      x^\ell
    \end{array}
    \right)
    =  \phi \left(
    \begin{array}{c}
      W_{11}^\ell z^{\ell-1} + W_{12}^\ell x^{\ell-1} \\
      W_{21}^\ell z^{\ell-1} + W_{22}^\ell x^{\ell-1}
    \end{array}
  \right).
\end{equation}

\vspace{-3mm}
\subsection{Internal feature separation into class likelihoods and contribution maps}
\vspace{-1mm}
The vector $z^\ell$ is expected to be a class likelihood at the $\ell$-th layer, especially in the case where $W^\ell$ is near the identity matrix. This is because $W_{11}^\ell \approx I_C$ and $W_{12}^\ell \approx O$ in this case, and thus $z^{\ell-1} \approx z^{\ell}$. However, it is also not true that $z^0 = 0 \approx z^L$, and thus there is a small increment from $z^{\ell-1}$ to $z^\ell$. This indicates that $W_{12}$ still has the power to extract some discriminative features from $x^{\ell-1}$ and adds it to $z^\ell$. The vector $z^\ell$ encodes the class likelihood at the $\ell$-th layer if $W^\ell$ is near the identity matrix as follows: First, $z^L$ is the class likelihood as noted above. Then, if $W^\ell$ is near the identity matrix, $z^{\ell-1}$ is almost the same as $z^{\ell}$ by \eqref{eq:forward_with_submatrix}. However, $z^0$ is a zero vector, and therefore the network should construct the class likelihood layer-by-layer by aggregating discriminative information from $x^{\ell-1}$ by multiplying $W_{12}^\ell$. Thus, $z^\ell$ is interpretable as the class likelihood at the $\ell$-th layer. \par

The variation from $x^0$ to $x^\ell$ is interpretable as the contribution map for classification. If $W^\ell$ is near the identity matrix, the output of each hidden layer is a slight variation of its input. The amount of variation is expected to be the contribution to classification because of the nature of an NN that gradually changes internal features to discriminative ones. 

\vspace{-3mm}
\section{Experiments}
\vspace{-2mm}
We conducted classification experiments to verify the validity of the proposed method. In this experiment, first, we evaluated the learning behavior to verify whether an identity-initialized MLP is trained without vanishing/exploding gradient and the learned weight matrices are close to the identity matrices. Second, we qualitatively evaluated the interpretability of the internal features by visualizing $z^\ell$ and $x^\ell$. Finally, we quantitatively evaluated the internal features.
\vspace{-3mm}
\subsection{Experimental conditions}
\vspace{-1mm}
We used the Fashion-MNIST and CIFAR-10 datasets. The parameters were set to $L=100$, $\sigma = 1 + \epsilon$, and $q^0 = q^* = 0.29$, where $\epsilon = 8 \times 10^{-4}$. We used hard-tanh functions as activation functions for the hidden layers and a softmax function for $z^L$ of the last layer. \par

\vspace{-3mm}
\subsection{Learning behavior of an identity-initialized MLP}
\vspace{-1mm}
\begin{figure}[t]
    \includegraphics[keepaspectratio,width=0.95\hsize]{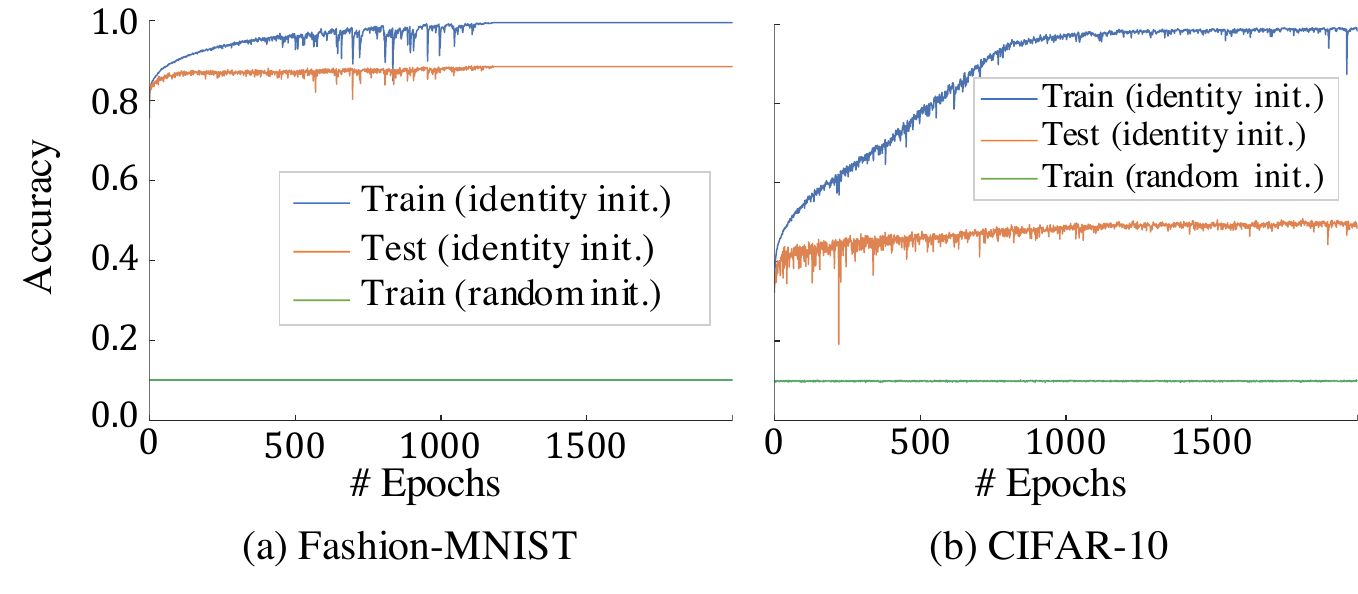}
  \caption{Classification accuracy for Fashion-MNIST and CIFAR-10.}\label{fig:learning_profile}
\end{figure}
Fig.~\ref{fig:learning_profile} shows the learning profile for each dataset. For comparison, the training results of CIFAR-10 using a 100-layer randomly initialized ReLU MLP are also shown. In both datasets, the training accuracy of the identity-initialized MLP monotonically increased and finally reached 1.0. These results demonstrate that identity-initialized MLPs can be trained without gradient vanishing/exploding occurrence as proved in \ref{sec:dynamical isometry}. In contrast, randomly initialized ReLU MLP training accuracy did not increase. The test accuracy is much lower than training because the theoretical result in \ref{sec:dynamical isometry} guarantees learnability, not generalization capability. However, it is almost the same level as a randomly initialized shallow MLP~\cite{zhang2016understanding}. \par

\begin{figure}[t]
    \centering
    \includegraphics[keepaspectratio,width=0.78 \hsize]{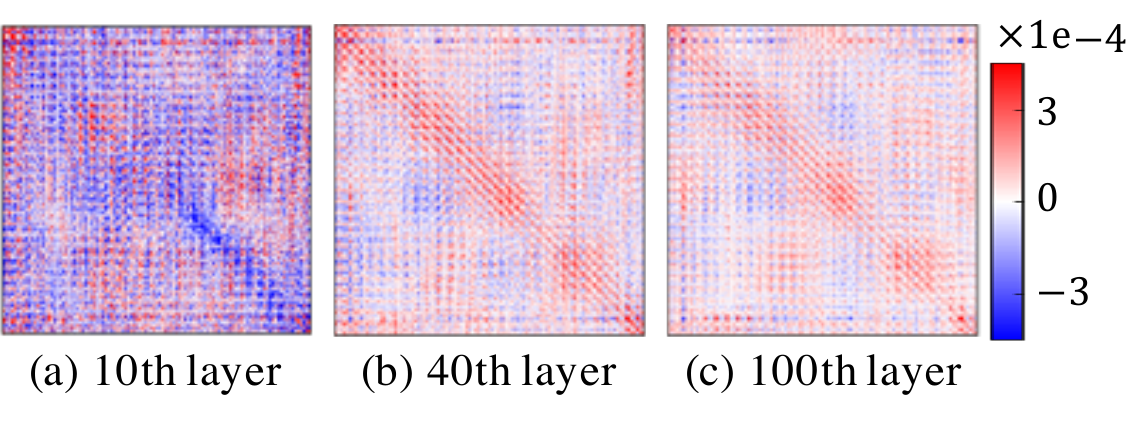}
  \caption{Difference of the learned weights from the initial identity matrix.}\label{fig:learned_weight}
\end{figure}
Fig.~\ref{fig:learned_weight} shows the variation of the hidden layers' learned weight matrices from the identity matrix for the Fashion-MNIST dataset. The results show that all the elements are remarkably smaller than 1. This demonstrates that the weight matrices of the hidden layers are near the identity matrices.\par

\vspace{-3mm}
\subsection{Qualitative evaluation}
\vspace{-1mm}
\begin{figure}[t]
  \centering
    \includegraphics[keepaspectratio,width=0.95 \hsize]{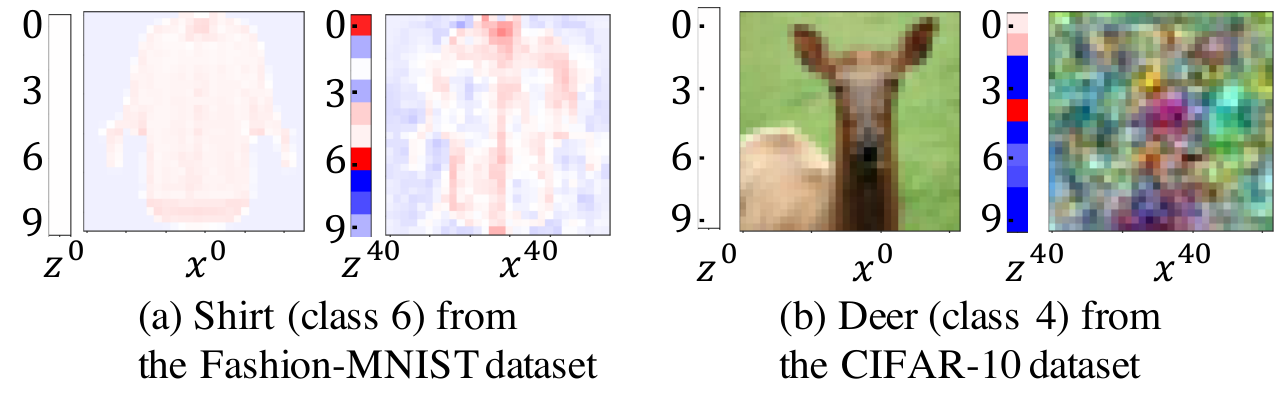}
  \caption{Input images and the related internal features.}\label{fig:shirt}
\end{figure}

Fig.~\ref{fig:shirt} shows example pairs of an input image and the related internal features at the 40th layer. In Fig.~\ref{fig:shirt}(a), for the input image belonging to class 6 (``shirt''), the hidden layer output $z^{40}$ had the highest value for the correct class 6 and the next highest for class 1 (``T-shirt''), which is a similar type of clothing to class 6. In $x^{40}$, the shoulders and sleeves, which are necessary information to classify ``shirt'' class, are emphasized. In Fig.~\ref{fig:shirt}(b), $z^{40}$ showed the highest value for the correct class (``deer''), and $x^{40}$ highlighted the face of the deer. These results suggest that $z^\ell$ involves class likelihood and $x^\ell$ represents the contribution to classification. \par

\vspace{-3mm}
\subsection{Quantitative evaluation}
\vspace{-1mm}
We conducted a qualitative evaluation of the internal features to reveal at which layer discriminative information appears. Here, we define an evaluation index of discriminability for the contribution map $x^\ell$. By ignoring the activation functions and assuming that the sub-matrices of $W^\ell$ are approximately regarded as $W_{11}^\ell = I_C$ and $W_{22}^\ell = I_N$, the forward propagation up to the $\ell$-th layer is approximated as
\begin{equation}
\phi \circ W^\ell \circ \cdots \circ \phi \circ W^1  
\!\approx\! \left(\!\!
    \begin{array}{cc}
      I_C & \sum_{k=1}^{\ell} W_{12}^k \\
      \sum_{k=1}^{\ell} W_{21}^k & I_N 
    \end{array}
  \!\!\right).
\end{equation}
As stated, $W_{12}^\ell$ conveys discriminative information from $x^{\ell-1}$ to $z^\ell$. In particular, the $c$-th row of $W_{12}^\ell$, $W_{12, c}^\ell$, extracts the information about class $c$ from $x^{\ell-1}$. Therefore, for the input data belonging to class $c$, if $x^\ell - x^0$ has a higher correlation with $\sum_{k=1}^\ell W_{12, c}^{k}$ than $\sum_{k=1}^\ell W_{12, c'}^{k}$, where $c' \neq c$, the variation of $x^\ell$ from the input contains discriminative information. We also calculated the classification accuracy for each layer by verifying the coincidence between the argmax of $z^\ell$ and the ground truth for the evaluation of $z^\ell$. \par

\begin{figure}[t]
    \centering
    \includegraphics[keepaspectratio,width=0.85\hsize]{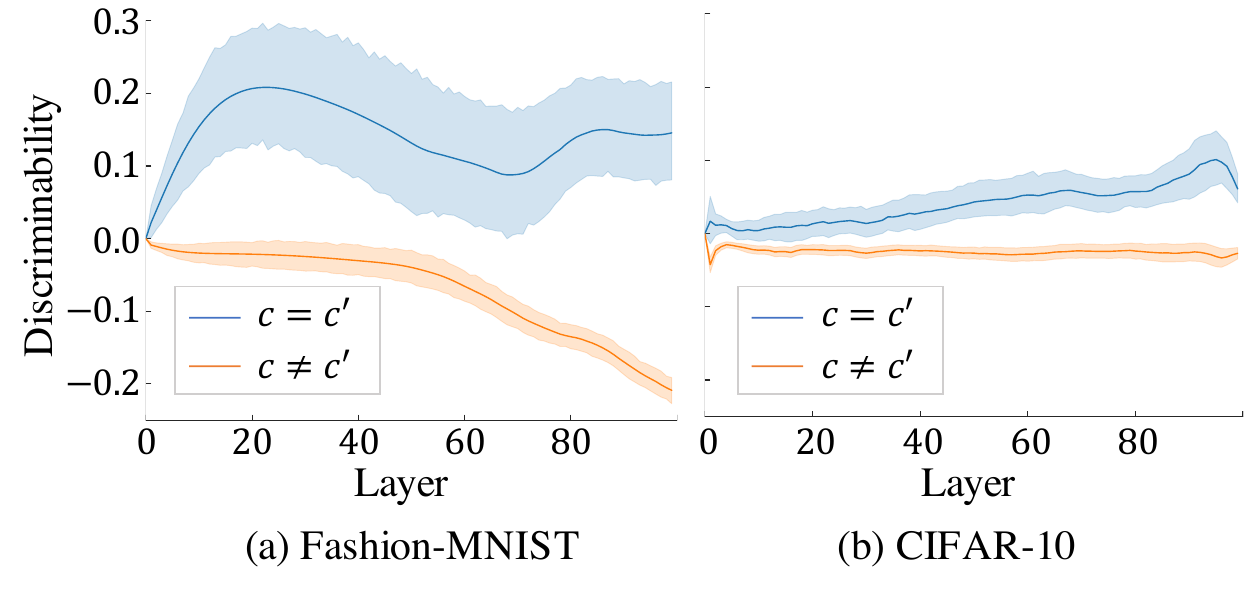}
  \caption{Layer-wise discriminability of the contribution map.}\label{fig:corr_mean}
\end{figure}
Fig.~\ref{fig:corr_mean} shows the layer-wise discriminability of the contribution map. These results indicate that the evaluation index demonstrates higher values when $c=c'$ than when $c \neq c'$. In particular, the discrepancy is larger in layers closer to the last layer, thereby showing that the network makes the contribution map more discriminative as it gets closer to the output. \par

\begin{figure}[t]
    \centering
    \includegraphics[keepaspectratio,width=0.85\hsize]{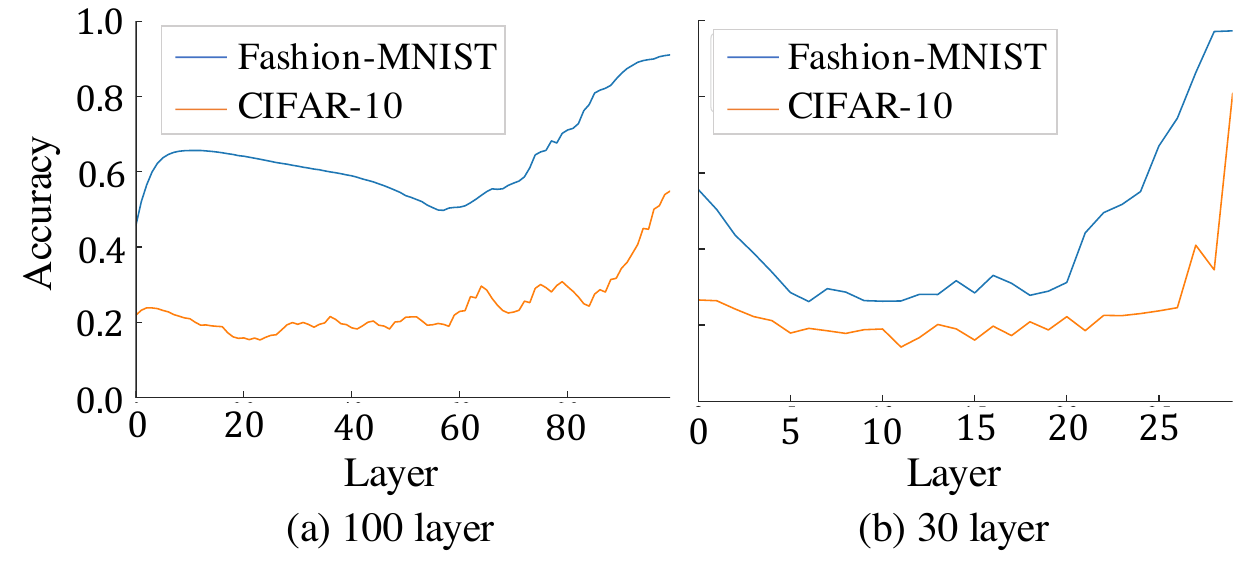}
  \caption{Layer-wise classification accuracy.}\label{fig:layerwise acc}
\end{figure}
Fig.~\ref{fig:layerwise acc} shows the layer-wise classification accuracy. For comparison, the results of a 30-layer MLP are also shown. The accuracy started to increase from approximately 30\% to 40\% of the layers close to the output without depending on the dataset and the number of layers. These results suggest that the first 70\% of the MLP is mainly responsible for feature extraction and the remaining 30\% for classification.

\vspace{-5mm}
\section{Conclusion}
\vspace{-3mm}
In this paper, we proposed an interpretation method of an MLP based on identity initialization. We first conducted a theoretical analysis of the identity-initialized MLP and showed that the forward/backward signal propagates without vanishing/exploding via dynamical isometry. We then proposed an interpretable MLP structure in which the features in each layer are divided into a contribution map and class likelihood, thereby allowing the quantification of the starting layer of classification. In our future work, we will investigate changes in contribution maps and the starting layer of classification during learning.
}

\bibliographystyle{IEEEbib}
\bibliography{refs}

\end{document}